\documentclass{article}

\usepackage[nonatbib,preprint]{neurips_2020}
\usepackage[utf8]{inputenc} 
\usepackage[T1]{fontenc}    
\usepackage[colorlinks=true,citecolor=red]{hyperref} 
\usepackage{url}            
\usepackage{booktabs}       
\usepackage{amsfonts}       
\usepackage{nicefrac}       
\usepackage{microtype}      
\usepackage{graphicx}
\usepackage{amsmath,bm}
\usepackage{amssymb}
\usepackage{array}
\usepackage{comment}
\usepackage{booktabs}
\usepackage{subfigure}
\usepackage{caption}
\usepackage{xcolor,colortbl} 
\usepackage{multirow}
\usepackage{makecell}
\usepackage{wrapfig}

\usepackage[ruled,vlined]{algorithm2e}
\usepackage[normalem]{ulem}

\DeclareMathOperator*{\argmax}{argmax}
\newcommand{\citep}[1]{\cite{#1}}
\newcommand{\citet}[1]{\cite{#1}}

\newcommand{\model}{\textsc{tgls}}
\newcommand{\syx}{s(\rm{y}|\rm{x})}
\newcommand{\brsup}[1]{^{(#1)}}
\makeatletter
\newcommand{\printfnsymbol}[1]{%
  \textsuperscript{\@fnsymbol{#1}}%
}
\title{Unsupervised Text Generation by\\ Learning from Search}

\author{
    Jingjing Li$^{1}$\thanks{equal contribution}\ \ ,~Zichao Li$^{2}$\printfnsymbol{1},~Lili Mou$^3$,~Xin Jiang$^2$,~Michael R. Lyu$^1$,~Irwin King$^1$\\
    $^1$The Chinese University of Hong Kong~~~~$^2$Huawei Noah's Ark Lab\\
    $^3$University of Alberta; Alberta Machine Intelligence Institute (Amii) \\
    \texttt{\{lijj,lyu,king\}@cse.cuhk.edu.hk}\\
    \texttt{\{li.zichao,jiang.xin\}@huawei.com}\\
    \texttt{doublepower.mou@gmail.com}\\
}
\begin{document}

\maketitle

\begin{abstract}
In this work, we present \model{}, a novel framework to unsupervised \textsc{t}ext \textsc{g}eneration by \textsc{l}earning from \textsc{s}earch. We start by applying a strong search algorithm (in particular, simulated annealing) towards a heuristically defined objective that (roughly) estimates the quality of sentences. Then, a conditional generative model learns from the search results, and meanwhile smooth out the noise of search. The alternation between search and learning can be repeated for performance bootstrapping. 
We demonstrate the effectiveness of \model\ on two real-world natural language generation tasks, paraphrase generation and text formalization. Our model significantly outperforms unsupervised baseline methods in both tasks. Especially, it achieves comparable performance with the state-of-the-art supervised methods in paraphrase generation.

\end{abstract}

\section{Introduction}\label{sec:intro}
Text generation refers to a wide range of tasks involving generating natural language, including but not limited to machine translation~\cite{lample2017unsupervised, lample2018phrase, lample2019cross}, sentence simplification~\cite{narayan2015unsupervised, surya2019unsupervised}, and text summarization~\cite{chu2018meansum, amplayo2020unsupervised}. Recent success of neural-based text generation  relies heavily on a large parallel dataset for training, which 
may not be available in real-world natural language processing (NLP) applications.
In this work, we consider unsupervised text generation, where no parallel data is available. This setting is more challenging, and has significant potential in both scientific research (e.g., low-resource language processing) and industrial applications (e.g., cold start for a new NLP application). 

Early work tackles unsupervised text generation by rules or templates~\cite{weizenbaum1966eliza, mcroy2003augmented}. While such approaches do not require parallel corpora, the generated sentences are highly subject to the rules, and hence lack the flexibility of natural language. Other work constructs pseudo-parallel data, which is only feasible for certain tasks like unsupervised machine translation~\cite{lample2017unsupervised}.

Recently, researchers have developed search-based techniques for unsupervised text generation~\cite{miao2019cgmh, simplification, schumann2020discrete,Liu2019UnsupervisedPB}, where a heuristically defined scoring function evaluates the quality of a sentence, involving language fluency, semantic compliance, and other task-specific aspects. Then, the algorithm performs word-level edits (such as word deletion, insertion, and replacement) to search towards a (possibly local) optimum of the scoring function. With a reasonably designed scoring function, such approaches are shown to be effective in a variety of applications like paraphrase generation~\cite{miao2019cgmh, Liu2019UnsupervisedPB}, sentence summarization~\cite{schumann2020discrete}, and text simplification~\cite{simplification}.

However, the search-based approach has two major drawbacks: 1) The inference efficiency is low. To obtain an output sentence, the search algorithm would perform a few hundred steps of local edits and re-evaluations. This could be considerably slower than an autoregressive decoder, which generates words sequentially. 2) The search could yield noisy results, since the scoring function is defined heuristically and the search is conducted locally in a discrete sentence space.

To this end, we propose a new framework for unsupervised \textsc{t}ext \textsc{g}eneration by \textsc{l}earning from \textsc{s}earch (\model{}), which contains a strong search module that explores the sentence space, as well as a learning module that learns from the search results.  
For the search module, we adopt the simulated annealing (SA) algorithm. At each step, SA proposes a local edit by a neural network, and then either accepts or rejects the proposal based on a heuristically defined scoring function. 
For learning, we employ two methods to train the conditional generative model, word-level cross-entropy loss and the sequence-level max-margin loss. Within \model{}, the search and learning can be boosted by each other in an iterative fashion. That is, the search results serve as the pseudo-reference for training the conditional generator, which in turn benefits SA search by serving as a more meaningful initial state.
As for implementation, \model{} involves two pretrained language models: a) the uni-directional GPT2~\cite{radford2019language}, which is suitable for likelihood-based fluency evaluation and conditional generation; and b) the bi-directional RoBERTa~\cite{liu2019roberta}, which is better at semantic evaluation and contextual word-level prediction.


The main contributions of our paper include: 1) We propose \model, a generic learning-from-search framework for unsupervised text generation. 2) We demonstrate efficient methods of incorporating the large-scale pretrained language models into our \model{} framework. 3) We conducted experiments on two different tasks: paraphrasing and text formalization. In both experiments, \model\ significantly outperforms unsupervised baseline methods. Moreover, \model\ achieves comparable performance to recent supervised models~\cite{du2019empirical} in the paraphrasing task. 4) For text formalization (an example of text style transfer), we are also the first to design a search-based 
method, and further extend it into the proposed \model\ framework.

\section{Approach}

\begin{figure*}[t!]
    \centering
    \includegraphics[width=1.0\textwidth]{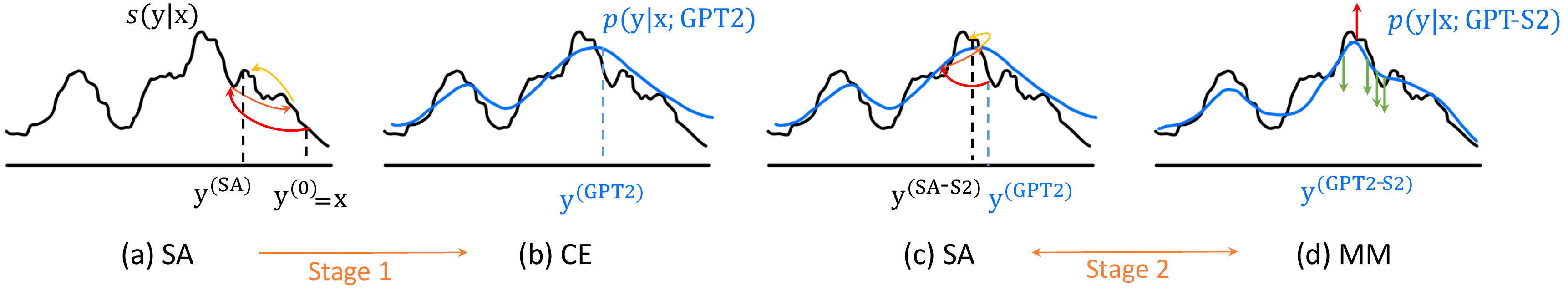}
    \caption{Overview of \model. (a) First-stage search by simulated anealing (SA). (b) First-stage learning by cross-entropy (CE) loss. (c) Second-stage search by SA. (d) Second-stage learning by max-margin (MM) loss. The horizontal axis represents the sentence space.}
    \label{fig:tgsl}
\end{figure*}

Our \model\ framework involves two stages of search and learning. In the first stage, we perform simulated annealing (SA) search~\cite{Liu2019UnsupervisedPB} and treat the obtained output sentences as pseudo-references. Then, we train an autoregressive GPT2 as the text generator~\cite{radford2019language} by word-level cross-entropy (CE) supervised learning, which enables our model to learn quickly. In the second stage, the search is conducted in a hybrid approach: the GPT2 executes beam search and the outputs are taken as the initial state of the SA algorithm again for iterative performance improvement. Later, we perform max-margin (MM) learning to better distinguish between higher-scored sentences and other high-probability but sub-optimal sentences. Figure~\ref{fig:tgsl} provides an overview of the two stages of search and learning in \model{}.

\subsection{Simulated Annealing Search}\label{sec:SA}
The search-based text generation~\cite{miao2019cgmh,Liu2019UnsupervisedPB} relies on a heuristic-based objective function $s(\rm y|\rm x)$ that (roughly) evaluates the quality of an output sequence $\rm y$ given the input $\rm x$ (usually, one or a few sentences). Typically, the objective involves language modeling fluency $s_\text{lm}(\rm x)$, semantic compliance $ s_{\text{semantic}}(\rm x, \rm y)$, and other task-specific scorers $s_\text{task}(\rm y,\cdot)$. 
These individual scorers are combined by the product of experts~\cite{PoE}:
\begin{align}\label{equ:obj}
 s(\mathrm{y}|\mathrm{x}) =
    s_{\text{lm}}(\mathrm{y}) \cdot s_{\text{semantic}}(\mathrm{x}, \mathrm{y}) \cdot s_{\text{task}}(\mathrm{y}, \cdot)
\end{align}

We adopt simulated annealing (SA)~\cite{kirkpatrick1983optimization,Liu2019UnsupervisedPB}, which performs local stochastic search to maximize the objective. Concretely, SA starts from an initial candidate output sentence $\mathrm y\brsup{0}$, which is set to the input $\mathrm x$ in our first-stage SA. For the second stage, it will be the output of our GPT2 model.

At a search step $t$, SA iteratively proposes a new candidate $\rm y'$ by local edits of $\mathrm y^{(t)}$, namely, word insertion, deletion, and replacement. The proposal $\rm y'$ is accepted with probability $p(\text{accept}|\mathrm y', \mathrm{y}\brsup{t}, \mathrm{x}, T) = \text{min}\big\{1, \exp(\frac{s(\mathrm y'|\mathrm x)-s(\mathrm y\brsup{t}|\mathrm x)}{T})\big\}$. Then, $\mathrm y\brsup{t+1}=\mathrm y'$ if $\rm y'$ is accepted, or otherwise, $y\brsup{t+1}=\mathrm y\brsup{t}$. In SA, $T$ is a temperature controlling how greedy the search algorithm is. Usually, $T$ is high at the beginning of search so as to be more explorative, and then $T$ is cooled down to achieve a better (local) optimum. 
Although we follow the generic SA framework of text generation as in~\cite{Liu2019UnsupervisedPB}, the objective function and proposal are largely redesigned, detailed below.

\textbf{Fluency scorer (}$s_\text{lm}$\textbf{).} The fluency of a sentence can oftentimes be approximated by a language model's predicted probability. Previous search-based work uses recurrent neural networks for fluency evaluation~\cite{miao2019cgmh,Liu2019UnsupervisedPB}. In our work, we use the large-scale pretrained GPT2 model~\cite{radford2019language}. For an output $\mathrm y = y_1\cdots y_n$, the language fluency scorer is the joint likelihood of $\mathrm y$, given by $s_{\text{lm}}(\mathrm y) =(\prod_{i=1}^{n} p(y_i|y_1, \cdots , y_{i-1}))^\alpha$, where $\alpha$ is a hyperparameter balancing $s_\text{lm}$ with other scorers in (\ref{equ:obj}). In fact, we use the vocabulary of GPT2 with bype-pair encoding (BPE), and $y_i$ here is a token after BPE segmentation. Our GPT2 is fine-tuned with non-parallel in-domain corpora to learn the specificity of a task. 

\textbf{Semantic scorer (}$s_\text{semantic}$\textbf{).} In this part, we extend the semantic scorers in~\cite{Liu2019UnsupervisedPB} with a  RoBERTa~\cite{liu2019roberta}. Fine-tuning details are presented in Appendex~\ref{app:roberta}. Compared with autoregressive GPT2 used for fluency evaluation, RoBERTa is pretrained by masked language modeling, and is better at feature representation. Let $\mathrm x=(x_1, \cdots, x_m)$ be a sentence. RoBERTa computes a contexualized representation of a word in the sentence as $\text{RoBERTa}(x_i, \mathrm x)$. 

A word-level semantic scorer evaluates how much keyword information (detected by Rake~\cite{rose2010automatic}) is preserved, given by the least matched keyword of $\mathrm x$:
\begin{equation}
    s_\text{word}(\mathrm y, \mathrm x)=\min_{k\in \operatorname{keyword}(\mathrm x)}\max_{y_i\in\mathrm y} \text{RoBERTa}(k, \mathrm x)^\top \text{RoBERTa}(y_i, \mathrm y)
\end{equation}

A sentence-level semantic scorer evaluates the cosine similarity of two sentence vectors $s_\text{sent}(\mathrm y,\mathrm x)=\frac{\bm y^\top \bm x}{\|\bm y\,\|\bm x\|}$, where the sentence vector is given by the RoBERTa feature of the padded token [BOS] at the beginning end of a sentence, i.e.,  $\bm x = \text{RoBERTa}(\text{[BOS]}, \mathrm x)$ and $\bm y$ is computed analogously.

Finally, the semantic scorer is the product of both word- and sentence-level scores as 
\begin{align}
    s_{\text{semantic}}(\rm y, \rm x) = s_{\text{word}}(\rm y, \rm x)^\beta \cdot s_{\text{sent}}(\rm y, \rm x)^\gamma
\end{align}
where $\beta$ and $\gamma$ are weighting hyperparameters.

\textbf{Task-specific scorers.} We apply \model\ to two tasks: paraphrasing and text formalization. 

For paraphrasing, the goal is to generate a semantically similar but lexically different sentence. Previous work~\cite{Liu2019UnsupervisedPB} uses the BLEU score to penalize the $n$-gram overlapping between the output and input: $s_\text{paraphrase}(\mathrm y, \mathrm x) = (1-\operatorname{BLEU(\mathrm y, \mathrm x))^{\delta}}$, which is also adopted in our work. Here, $\delta$ is a weighting hyperparameter for the task-specific scorer.

For text formalization, the goal is to transform an informal sentence to the formal style~\cite{Rao2018DearSO}, which is an example of text style transfer. We follow the setting of most text style-transfer work~\cite{hu2017toward}, where we assume the style labels are available, but no parallel supervision is given. 
We train a classifier that predicts the probability of the style, also based on the RoBERTa features. Then, the task-specific scorer becomes  $s_\text{formality}(\mathrm y) = ( p(\text{formal}\,|\,\text{RoBERTa}(\text{[BOS]},\mathrm y)))^{\delta}$, 
where $\delta$ is the weighting hyparaparameter for this task.

\textbf{Proposal of local edits.} At a step $t$ of SA search, a new candidate $\mathrm y'$ is proposed from $\mathrm y\brsup{t}$ by local editing. SA randomly picks a position to edit, as well as one of the following edits: \texttt{Replace}, \texttt{Insert}, and \texttt{Delete}.

For \texttt{Replace}, the model suggests a candidate word at $x_i$ based on the posterior distribution induced by $\syx$. For efficiency concerns, previous work~\cite{miao2019cgmh,Liu2019UnsupervisedPB}  evaluates top-$K$ candidate words, suggested by a forward and backward language model. In our work, we adopt RoBERTa to evaluate the posterior probability of a word, where the word embedding layer of RoBERTa at this slot is randomly masked. 
The \texttt{Insert} edit also suggests a word from the posterior, predicting a word given the newly added [MASK] token and the context. This complies with RoBERTa's pretraining criteria of masked language modeling and is able to suggest high-quality candidate words. The \texttt{Delete} operator simply removes the word at a chosen position.

In text formalization, we have local edits based on a set of rules, e.g., ``\textit{we are}'' substituting ``\textit{we're}.'', which are retrieved from PPDB\cite{pavlick2015ppdb}. Previous sequence-to-sequence approaches on this task adopt manually designed rules as a preprocessing step~\cite{Rao2018DearSO}. Our unsupervised \model{}, on the other hand, can easily make use of the off-the-shelf resources, since it can filter out the noise by rejecting the bad candidates.

In short, the SA search component in our \model\ mainly follows~\cite{Liu2019UnsupervisedPB}, but we re-design the scoring functions and the proposals. The main focus of this paper is to couple search and learning, especially the methods of training a machine learning model that learns from the search results, as follows.

\subsection{Word-Level Cross-Entropy (CE) Learning}\label{sec:sbs}

As mentioned in Section~\ref{sec:intro}, the local search algorithm is computationally inefficient during inference time, because it requires a few hundred steps of edits and re-evaluations for each sample.

Our intuition is to train a conditional generative model, GPT2, based on SA's search results. Specifically, an input $\rm x$ and SA's searched sequence $\rm y\brsup{\text{SA}}$ are concatenated with a special separating token [SEP] in between, and the GPT2 is trained with losses on the $\rm y$-part.  Therefore, the GPT2 would be able to generate a output sequence directly from $p(\mathrm y|\mathrm x)$ in an autoregressive way.

Given a source sequence $\rm x$, the objective is the word-by-word cross-entropy (CE) loss, given by
\begin{equation}\label{eqn:sbs}
    J_\text{CE} = -\sum_{n=1}^N \sum_{v\in\mathcal{V}}  y^{(\text{SA})}_{i,v} \log p^{(\text{GPT2})}_{i,v}
\end{equation}
where $ y_{i,v}^{(\text{SA})}$ is a binary value, indicating whether the $i$th word is $v$ or not in the SA's output for this data sample, and   
$p^{(\text{GPT2})}_{i,v}=\Pr\big [y_i=v\,|\, {\mathrm y}^{(\text{SA})}_{< i}, \mathrm x\big]$, which is predicted by the GPT2. 

The word-level CE learning in \model\ adopts standard teacher-forcing technique with SA's output being the pseudo-reference, i.e., during training, the GPT2 model learns the probability $p^{(\text{GPT2})}_{i,v}$ at step~$i$, assuming all previous words are correctly predicted as $ {\mathrm y}^{(\text{SA})}_{< i}$. Thus, word-by-word CE trains all predictions in the sequence simultaneously, and is able to quickly adapt a generic pretrained GPT2 to the text generation task at hand.  

It is also noted that minimizing the cross-entropy loss~(\ref{eqn:sbs}) is equivalent to minimizing $\operatorname{KL}(\widehat{\bm y}_i^{(\text{SA})}\|{\bm p}_i^{(\text{GPT2})})$, i.e.,  the KL-divergence between $\widehat{\bm y}_i^{(\text{SA})}$ and ${\bm p}_i^{(\text{GPT2})}$, if viewed as distributions over the vocabulary. Due to the asymmetry nature, minimizing the KL-term makes the second slot ${\bm p}_i^{(\text{GPT2})}$ more wide-spreading than the first slot $\widehat{\bm y}_i^{(\text{SA})}$, illustrated in Figure~\ref{fig:tgsl}b. This provides an explanation of why the CE-trained GPT2 could smooth out the noise of the stochastic SA search. As will be shown in experiments, training a GPT2 from SA's output alone can outperform SA. 

\subsection{Sequence-Level Maximum-Margin (MM) Learning}
Our next insight is to perform alternations between search and learning to bootstrap performance of \model{}. 
In the first stage, SA starts local search with the initial candidate being the input (i.e., $\mathrm y\brsup{0}=\mathrm x$), because we have no other meaningful candidate output yet. Starting with $\mathrm x$ takes advantage of the resemblance between input and output. But if a higher-quality candidate is available, SA may perform better than from $\rm x$.

Therefore, we propose to have another stage of learning-from-search alternations.
SA starts from an initial candidate being GPT2's output, i.e., $\mathrm y\brsup{0}={\rm y}\brsup{\text{GPT2}}$, shown in Figure~\ref{fig:tgsl}c.
Then, GPT2 is further fine-tuned to learn from the newly searched result. For the learning method, we propose to employ sequence-level max-margin (MM) training, instead of CE training. Such alternation can be performed for multiple epochs for performance bootstrapping. 

Concretely, the GPT2 trained with CE learning performs beam search (beam size $B$) and obtain a set of output sequences $Y\brsup{\text{GPT2}}=\{\mathrm y\brsup{\text{GPT2}, 1}, \cdots, \mathrm y\brsup{\text{GPT2}, B}\}$. A randomly picked (for efficiency purpose) output in $Y\brsup{\text{GPT2}}$ is taken as initial candidate in SA search, yielding a new sample $\mathrm y^{(\text{SA-S}2)}$. We consider the set $\widetilde Y=Y\brsup{\text{GPT2}}\cup\{\mathrm y^{(\text{SA-S}2)}\}$ as the positive and negative samples for MM learning.
In fact, the positive sample $\rm y^+$ is the best sequence scored by (\ref{equ:obj}), i.e., $\mathrm y^+=\operatorname{\argmax}_{\mathrm y\in\widetilde Y} \syx$. In most cases, we have $\mathrm y^+=\mathrm y^{(\text{SA-S}2)}$, but this is not necessarily true because SA is not greedy. All other sentences in $\widetilde Y$ are collected as negative samples. We use the summation of GPT2's pre-softmax logit as the negative energy.\footnote{\textit{Energy} is what MM learning would like to minimize for positive samples.} In other words, we have $-E(\mathrm y)=\sum_{i=1}^N z_{i,y_i}$ of a sequence $\mathrm y=(y_1, \cdots, y_N)$, where $z_{i,y_i}$ is the logit for the word $y_i$ at the $i$th step. The max-margin loss for this data sample is
\begin{equation}\label{eqn:MM}
J_\text{MM}=\sum_{\mathrm y^-\in\widetilde Y,\, \mathrm y^-\ne \mathrm  y^+}\max\big\{0, -E(\mathrm  y^+) + E(\mathrm y^-) -\Delta\big\}
\end{equation}
where $\Delta$ is the margin hyperparameter.

In fact, the energy implicitly defines a globally normalized distribution as $p(\mathrm y)=\frac1Z \exp\{-E(\mathrm y)\}$, where $Z$ is the partition function. The MM training increases the probability of the positive sample, while decreasing the probability of negative ones. 
In our MM training, the negative samples are given by beam search on GPT2, highly resembling the positive one. This makes \model\ more sensitive to the sequence-level scorer~(\ref{equ:obj}) in its probable region of the output space, illustrated in Figure~\ref{fig:tgsl}d. 

By contrast, word-level CE increases the probability of the target (analogous to the positive sample) step-by-step, while decreasing the probability of other samples due to local normalization. Thus, it cannot explicitly correct the prediction of a highly-probable but low-scored sample, and performs worse than MM in the second stage.

\begin{algorithm}[t]
\small
 \textbf{Input:} A non-parallel corpus $X$\\
 \textbf{Output:} A fine-tuned GPT2 model \\

 {\color{blue}$\rhd$ First-stage search and learning}\\
 \For{an input $\mathrm x\in X$ }{
    $\mathrm y\brsup{\text{SA}}=\operatorname{SA}(\mathrm x, \mathrm x)$\\ \quad\quad{\color{gray}$\rhd$ SA is detailed in Algorithm~\ref{alg:SA}. In the first stage, SA starts with input $\rm x$ as the initial candidate}
 }
 \For{all epochs}{
    \For{an input $\mathrm x$ with its SA output $\mathrm y\brsup{\text{SA}}$}{
    Fine-tune GPT2 by cross-entropy loss (\ref{eqn:sbs}) with pseudo-reference $\mathrm y\brsup{\text{SA}}$, conditioned on $\rm x$
    }
}
  {\color{blue}$\rhd$ Second-stage search and learning}\\
 \For{all epochs}{
    \For{an input $\rm x$}{
        
        $Y\brsup{\text{GPT2}}=\operatorname{BeamSearch}(\text{GPT2}( $\rm x$))$\quad  {\color{gray}$\rhd$ $Y\brsup{\text{GPT2}}$ is a set of output by beam search}  \\[3pt]
        $\mathrm y\brsup{\text{SA-S2}}=\operatorname{SA}(\mathrm x, \mathrm y\brsup{\text{GPT2}})$ for some $\mathrm y\brsup{\text{GPT2}}\in Y\brsup{\text{GPT2}}$\\
        \quad\quad{\color{gray}$\rhd$ In the second stage, SA starts with GPT2's output (any output in the beam is fine)}\\[3pt]
        $\widetilde Y= Y\brsup{\text{GPT2}}\cup \{\mathrm y\brsup{\text{SA-S2}}\}$\\[3pt]

        Fine-tune GPT2 with max-margin loss (\ref{eqn:MM}) with \\[2pt]
\quad\quad positive sample:                $\mathrm y^+=\argmax_{\mathrm y\in \widetilde Y} s(\mathrm y|\mathrm x)$, and\\
\quad\quad negative samples: $\widetilde Y\backslash\{\rm y^+\}$\\
    }
 }
 \textbf{Return: }  Resulting GPT2 (denoted by GPT2-S2 after two stages of search and learning)\\

 \caption{Training \model}\label{alg:tgsl}
\end{algorithm}
In summary, the training of \model\ involves two stages of search and learning, where CE and MM are used as the learning objective in different stages. Notice that, for the second stage, search and learning are alternated within the epoch loop. Thus, another stage of search and learning is unnecessary, because our second stage already allows multiple epochs for performance bootstrapping.
For inference, we do not perform SA search, but directly use the fine-tuned GPT2 for autoregressive prediction.

Appendix~\ref{app:algo} further provides a detailed diagram of our \model{}.

\subsection{Discussion: \model{}~vs.~Reinforcement Learning and Learning-to-Search}
\label{sec:diss}

One of the most popular algorithms of reinforcement learning (RL) in text generation is the REINFORCE, which maximizes the expected reward (such as the BLEU score~\cite{seq-level} or an adversarial discriminator~\cite{seqGAN}) by sampling a sequence of actions from its learned policy and reweighing the likelihood training of the sampled sequence. REINFORCE is known to have high variance, and previous REINFORCE-based text generation involves groundtruth pretraining~\cite{seqGAN}. 
Without a warm start, the sampling-based REINFORCE does not work with such a large action space as the vocabulary. Our \model\ would also optimize an external scoring function (analogous to the reward in RL), but does not have grountruth for pretraining. We instead perform SA search and learn from SA's (local) optima step-by-step.


Monte-Carlo Tree Search (MCTS)~\cite{alphaGo} is another paradigm of search and learning, where a search tree is maintained with each non-leaf node being a partial configuration (e.g., a partial sentence in text generation). Again, it suffers from the large branching factor, which is the vocabulary size in our applications. Our \model\ adopts local search, which maintains a full configuration and evaluates the candidate at each search step. The resemblance 
between input and output also largely eases the search task. 

The \textsc{L}earning-to-\textsc{S}earch (\textsc{l2s}) framework has been successfully applied to various NLP applications, such as structured prediction~\cite{daume2005learning, krishnamurthy2015learning} and text generation~\cite{BSO, zhang2019bridging}. \textsc{l2s} allows the model to explore/search in the space, collects the score (cost) for possible actions, and optimizes the model. Usually, \textsc{l2s} assumes an expert demonstration (groundtruth sequence and/or dynamic oracle) available as a reference policy \textsc{l2s}. For instance, a LaSO-like algorithm forces the model to search towards the groundtruth sequence; when the groundtruth is out of the search range, a learning update is performed, where the search effort serves as the negative samples and the groundtruth as positive examples for learning~\cite{LaSO,BSO}. By contrast, \model{} does not have groundtruth, but uses a strong search algorithm to find higher-scored sentences, which serve as positive samples.

\section{Experiments}
\subsection{Datasets and Settings}

\textbf{Paraphrase Generation.}
Paraphrase generation is to rephrase input text with different expressions, while keeping the semantics.
Following previous work~\citep{fu2019paraphrase, gupta2018deep}, we conducted experiments on the Quora benchmark dataset.\footnote{https://www.kaggle.com/c/quora-question-pairs} As unsupervised text generation, we followed \citep{Liu2019UnsupervisedPB} and used 500K sentences to fine-tune GPT2 and RoBERTa for fluency and semantic scorers. For validation and testing, we had 500 and 170K samples, respectively.

We adopt BLEU and iBLEU as evaluation metrics, which are widely used for paraphrase generation. BLEU measures the length-penalized $n$-gram overlap between an output and the reference.
In addition, paraphrasing requires that the output should be different from input. Thus, iBLEU~\cite{sun2012joint} penalizes BLEU by $n$-gram similarity between output and input. Following most work, we consider iBLEU as the main metric for paraphrasing.

\textbf{Text Formalization.}
This task concerns formality transfer of text, and our goal is to rephrase the given informal text into the formal style. 
We experimented with the Grammarly’s Yahoo Answers Formality Corpus (GYAFC) \citep{Rao2018DearSO} in the domain of Family \& Relationships. It is noted that GYAFC contains 50K informal--formal pairs, but our \model\ follows the setting of most other style-transfer work~\cite{hu2017toward}, which uses non-parallel corpora with style labels, but does not have parallel supervision. Our pretrained language models are additionally fine-tuned on automatically labeled non-parallel corpus \citep{Xu2019FormalityST}. In GYAFC, there are 3K samples for validation and 1K for test.

The performance of formality transfer is measured in different aspects. The language modeling perplexity evaluates the fluency of the generated text, and a learnable classifier predicts the formality accuracy. Particularly, the formality evaluator achieves an accuracy of 94\%, being a good automatic evaluation measure.\footnote{We reuse the architecture of RoBERTa for formality evaluation and GPT2 for fluency evaluation. However, they are separately trained, third-party models, and are NOT part of our \model.} The BLEU score is also computed against the reference to evaluate $n$-gram overlap. Finally, we consider the harmonic mean (H-mean) and the geometric mean (G-mean) of the formality accuracy and the BLEU score as our main metrics for this task. 

\textbf{Hyperparameters.} For SA, the initial temperature was set to 1e-2 in both tasks. The total search steps and temperature cooling were 50, 2e-4 for paraphrasing; and 100 and 1e-4 for text simplification.
The scorers' weights were tuned by grid search, set as ($\alpha, \beta, \gamma, \delta)= (0.8, 1, 0.6, 0.125)$ for paraphrasing, and $(0.8, 2, 1.25, 0.26)$ for text formalization. We keep the RoBERTa fixed and further tune the GPT2 model by alternations of searching-and-learning for another 6 epochs.


\begin{table}[t!]
    \begin{minipage}[t]{0.41\linewidth}
    	\caption{Automatic evaluation results on paraphrasing.}
    	\centering
    	\label{tab:auto_para_results}
    	\small
    	\resizebox{\linewidth}{!}{
    	\begin{tabular}{lcc}
    		\toprule
    		Methods & iBLEU & BLEU    \\ \midrule
    		\multicolumn{3}{c}{Supervised}\\ \midrule
	RL-\textsc{NN}~\cite{qian2019exploring} & 14.83 & 20.98 \\ 
    		\textsc{Dagger}$^\dag$~\cite{du2019empirical} & 18.88 & 28.42 \\
    		GPT2$^\dag$~\cite{radford2019language} & 19.19 & 26.92 \\ \midrule
    		\multicolumn{3}{c}{Distant supervised}\\ \midrule
    		Round-Trip MT (GPT2)$^\dag$~\cite{guo2019zero}& 11.24 & 16.33\\ 
    		Round-Trip MT (Transformer)$^\dag$~\cite{mallinson2017paraphrasing} & 14.36 & 20.85 \\ \midrule
    		\multicolumn{3}{c}{Unsupervised}\\ \midrule

    		VAE \citep{bowman2015generating} 
    		&  8.16 & 13.96 \\ 
    		CGMH \citep{miao2019cgmh}
    		&  9.94 & 15.73 \\ 
    		UPSA \citep{Liu2019UnsupervisedPB}
    		& 12.02 & 18.18   \\ 
    		SA w/ PLM (Ours)$^\dag$\!\! & 14.52 & 21.08\\
    		\model{} (Ours)$^\dag$
    		& \textbf{17.48} & \textbf{25.00}  \\ \bottomrule
    	\end{tabular}
    	}
    \end{minipage}
    \begin{minipage}[t]{0.01\linewidth}\ 
    \end{minipage}
\begin{minipage}[t]{0.59\linewidth}
\caption{Automatic evaluation results on formality transfer. $^\downarrow$The smaller, the better.}
    	\label{tab:auto_style_results}
    	\resizebox{\linewidth}{!}{
    	\begin{tabular}{l|ccc|cc}
    		\toprule
    		Methods$^\dag$  & PPL$^\downarrow$ & BLEU  & Formality & H-mean & G-mean   \\ \midrule
    		\multicolumn{6}{c}{Supervised}\\ \midrule
     		LSTM-attn \citep{Rao2018DearSO} 
     		& \textbf{23.42} & \textbf{69.36} & \textbf{87.39} & \textbf{77.34} & \textbf{77.85} \\
     		\midrule
    		\multicolumn{6}{c}{Unsupervised}\\ \midrule
    		BackTrans \citep{prabhumoye2018style} 
    		& 183.7 & 1.23 &  31.18 &  2.37 & 6.13 \\ 
    		StyleEmb \citep{fu2018style}
    		&	114.6 & 8.14 & 12.31 &   9.80 &	10.01 \\ 
    		MultiDec \citep{fu2018style}
    		& 187.2 &  13.29 & 8.18	 & 10.13 & 10.42  \\ 
    		CrossAlign \citep{shen2017style} 
    		& 44.78 & 3.34 & 67.34 & 6.36 & 14.99 \\ 
    	    DelRetrGen \citep{li2018delete}
    		& 88.52 & 24.95 & 56.96 & 34.70  & 37.69  \\ 
    		Template \citep{li2018delete} 
    		& 197.5 & 43.45 & 37.09 & 40.02  & 40.14  \\ 
    		UnsupMT \citep{zhang2018style}
    		& 55.16 & 39.28 & 66.29  & 49.33  & 51.02 \\ 
    		DualRL \citep{Luo19DualRL}
    		 & 66.96  & 54.18 & 58.26 & 56.15  & 56.18  \\  
    		\model{} (Ours)                     
    		& \textbf{30.26} & \textbf{60.25} & \textbf{75.15} & \textbf{66.88} & \textbf{67.29}  \\
    		\bottomrule
    	\end{tabular}
    	}

\small    	
	$\dag$ indicates that the results are directly comparable to \model\ on the same data split. Appendix~\ref{app:baseline} provides more details on the baseline models and how these results are obtained.

    \end{minipage}
\end{table}

\subsection{Overall Performance}\label{sec:auto}
Table \ref{tab:auto_para_results} presents the results of automatic evaluation for paraphrase generation. 
Among the unsupervised approaches, the simulated annealing model UPSA~\cite{Liu2019UnsupervisedPB} achieves the previous state-of-the-art performance, outperforming both variational sampling~\cite{bowman2015generating} and discrete-space Metropolis--Hastings sampling~\cite{miao2019cgmh}. We propose to use large-scale pretrained language models for fluency and evaluation (mode name: SA w/ PLM), and improve iBLEU by 2.5 points from UPSA. Our \model\ framework of search and learning further improves iBLEU by 2.96 points, being a new state-of-the-art unsupervised paraphrasing model.

The \model{} also outperforms the paraphrasing systems based on round-trip translation, which is widely used in real-world applications. Such methods generate a paraphrase by translating a sentence to a foreign language and translating it back. It is categorized as distant supervision, because it requires parallel corpora for machine translation, but not for the paraphrasing task of interest.

Noticeably, our unsupervised \model\ performs comparably to a few recent paraphrasing model~\cite{qian2019exploring,du2019empirical}.  Moreover, we train a GPT2 in the supervised setting for a controlled experiment, where the neural architecture is fixed. We see that the unsupervised \model\ is slightly worse than the supervised setting by only 1.71 iBLEU, largely closing the gap between supervised and unsupervised paraphrasing.

Table \ref{tab:auto_style_results} presents the results for formality transfer. Again, we see consistent evidence on the effectiveness of \model, as it outperforms existing unsupervised approaches including heuristic marking of style words and retrieval-based editing~\cite{li2018delete}, unsupervised machine translation approaches~\cite{zhang2018style}, and dual reinforcement learning~\cite{Luo19DualRL}. 

Admittedly, the unsupervised \model\ is still worse than supervised approaches on this task. This is probably because our heuristic scorers are mostly designed for the paraphrasing task, and even for large-scale pretrained models, their performance may drop with informal text. 
More effort could be made here for future work. 

We also conducted human evaluation, reported in Appendix~\ref{app:huamn}. Results are consistent with these automatic metrics.
\vspace{-5pt}
\subsection{Analysis}
In this part, we present an in-depth analysis of our model with paraphrase generation as the testbed.

\textbf{Ablation study.} As \model\ involves two stages of search and learning, we conduct an ablation study, shown in Table~\ref{tab:abl_para_results}. We start from a base simulated annealing (SA) approach, where we have already adopted pretrained language models. Thus, it sets up a fair comparison.

In the first stage of learning, our GPT2 model with word-level cross-entropy (CE) training already outperforms SA alone. The result is slightly surprising, but it actually makes sense because cross-entropy loss can smooth out the noise in SA's heuristically defined search objective. 

We also tried to train the GPT2 by max-margin (MM) loss without CE learning, but it fails to escape from a random policy. It is due to the difficulty of training an energy-based model in comparison to a locally normalized model~\cite{LN2BN}. In our work, the negative samples in the beam would be useless when the model is not warm started.

We compare SA with the initial sentence being input and GPT2's prediction (SA vs.~SA$+$CE$+$SA). We see the latter outperforms both SA and SA$+$CE. This confirms that the learned GPT2 helps SA find a better optimum.

The last two lines of Table~\ref{tab:analysis} provide evidence of performance bootstrap by alternating between search and learning, as they outperform other ablated variants. In particular, MM is better than CE by a significant margin in the second stage. Our intuition is that MM with negative samples in the beam makes \model\ more sensitive in distinguishing sentence quality with its highly probable output region.

\begin{table}[!t]
    	\caption{Model analysis on paraphrase generation. All variants use pretrained language models.}\label{tab:analysis}
    	\centering\small
    	\label{tab:abl_para_results}
    	\resizebox{!}{!}{
    	\begin{tabular}{l|cc|c}
    		\toprule Methods &\!\! iBLEU\!\! &\!\! BLEU &\!\!\!\! Inference Time\\
    		&&&($\text{sec}/\text{sample}$)   \\ \midrule
    		SA & 14.52 & 21.08  &  5.46 \\ 
    		SA$+$CE & 14.97 & 23.25 & 0.06 \\ 
    		SA$+$CE$+$SA & 15.41 & 21.48 & 2.62 \\
    		SA$+$CE$+$SA$+$CE & 15.70 & 21.70 & 0.37\\ 
    		SA$+$CE$+$SA$+$MM (full) \!\!   	& \textbf{17.48} & \textbf{25.00}  & 0.43\\           
    		\bottomrule
    	\end{tabular}}
 \end{table}

\textbf{Inference efficiency.} We also report computational time in Table~\ref{tab:analysis}. 
The experiments were conducted on a cluster with Nvidia Telsa V100 GPUs. The inference time could be noisy due to the multi-thread nature of clusters, but it provides a conclusive enough comparison between search-based and autoregressive generation.  As seen, SA is inefficient because it requires hundreds of steps of editing and reevaluation.
 SA$+$CE, SA$+$CE$+$SA$+$CE, and SA$+$CE$+$SA$+$CE are all based on the GPT2 model during inference, and thus are much more computationally efficient. Based on the validation, SA$+$CE adopts greedy decoding, whereas the other two adopt beam search with a size of $5$. We see all GPT2-based generators are at least 6--10$\times$ faster than the search-based methods.

\textbf{Case Study.} We present a case study in Appendix~\ref{tab:case}. Typically examples show that \model\ is able to generate more fluent and more different-appearing paraphrases than search-based methods.

\section{Related Work}

\textbf{Unsupervised text generation}. There has been extensive work on neural unsupervised text generation. One popular approach is the variational auto-encoder~\cite{kingma2013auto}, which is able to generate text by manipulated latent space for certain attributes, such as sentiment~\cite{hu2017toward}, topic~\cite{wang2019topic}, and syntax~\cite{zhang2019syntax}. 

Recently, search-based methods has been developed for various text generation tasks, including text simplification~\cite{simplification}, summarization\cite{schumann2020discrete}, keyword-to-text generation~\cite{miao2019cgmh}, and paraphrasing~\cite{miao2019cgmh, Liu2019UnsupervisedPB}. However, these methods are not learnable; hence, the inference is inefficient and the performance cannot be improved by training. 

Most of other work of unsupervised text generation is built upon heuristics of certain tasks. For instance, Narayan and Gardent~\cite{narayan2015unsupervised} propose a task-specific pipeline for sentence simplification. Zheng and Lapata~\cite{zheng2019sentence} employ a graph-based ranking algorithm to select the most significant sentence as the summarization of a document. Chu and Liu~\cite{chu2018meansum} utilize the overlapping of text as a hint for multi-document summarization. In our work, the proposed \model{} is a generic framework, which also allows encoding  prior knowledge of a task into the search algorithm.

\textbf{Paraphrase generation.} 
Most of the recent work on paraphrase generation focuses on neural models trained with large scale parallel-datasets. Several of them adopt different search techniques in supervised learning. The RL-based approaches propose to learn reward functions to score sampled sequence~\cite{li2017paraphrase, qian2019exploring, yang2019end}. Besides, Du and Ji~\cite{du2019empirical} finds that the learning-to-search approach can improve the performance of neural paraphrasing model. While our work mainly considers search and learning for unsupervised generation.

There is a group of work that utilizes (statistical- or neural-) machine translation (MT) systems, generating paraphrase by translating source sentences into a pivot language, and then translate it back into the original language~\cite{zhao2010leveraging, mallinson2017paraphrasing, guo2019zero}. Although no supervision of paraphrase is needed, the success of this approach depends on one or more high-quality MT systems, and hence large-scale parallel datasets for training translation models.

For unsupervised paraphrasing, one direction is to perform sampling by either a latent-space variational posterior sampler~\cite{bao2019generating} or a word-space Metropolis--Hastings (MH) sampler~\cite{miao2019cgmh}. By decreasing the temperature of the stationary distribution, Liu et al.~\cite{Liu2019UnsupervisedPB} show that search-based formulation outperforms sampling for unsupervised text generation. Our work further extends it to a learning-from-search framework, improving both accuracy and inference efficiency.

\textbf{Text style transfer.} Our text formalization is one application of text style transfer. Other examples include sentiment~\cite{hu2017toward} and the prose style~\citep{xu2012paraphrasing}.
Typically, text style transfer can be divided into three categories: parallel supervised, non-parallel supervised (with only style labels), and purely unsupervised. Parallel supervised style transfer trains a sequence-to-sequence model~\cite{Rao2018DearSO}, whereas purely unsupervised style transfer replies on disentangling latent space~\cite{unsupervised}. 

Most previous work on text style transfer is in the non-parallel supervised setting, assuming style labels are available. Researchers have developed style embedding-based approaches~\cite{shen2017style,fu2018style}, style-specific decoders~\cite{fu2018style}, style-word editing approaches~\cite{li2018delete}, among others. Our approach also follows this setting, but to the best of our knowledge, we are the first to model style transfer as a search problem, as well as to extend it to the proposed  \model\ framework of search and learning.

\vspace{-5pt}
\section{Conclusion}
\vspace{-5pt}
This work proposes \model{}, a novel framework of learning-from-search to unsupervised text generation. We show that the simulated annealing search can provide high-quality examples for the conditional text generator to learn from. Further, the improved generative model can give a better initial state to the search algorithm. Experiments demonstrate that the alternation of search and learning can boost the performance of \model{} on two unsupervised text generation tasks, paraphrase generation and text formalization. Moreover, our model is considerably more computationally efficient, compared with search-based generation methods. We note that \model{} opens a few future directions, such as more effective and efficient search algorithms, more noise-robust learning methods, and a better combination of search and learning.
We would also like to apply the learning-from-search framework to other sequential prediction tasks in NLP.

\bibliography{ref}
\bibliographystyle{abbrv}

\newpage
\appendix
\begin{figure}[!t]
    \centering
    \includegraphics[width=\linewidth]{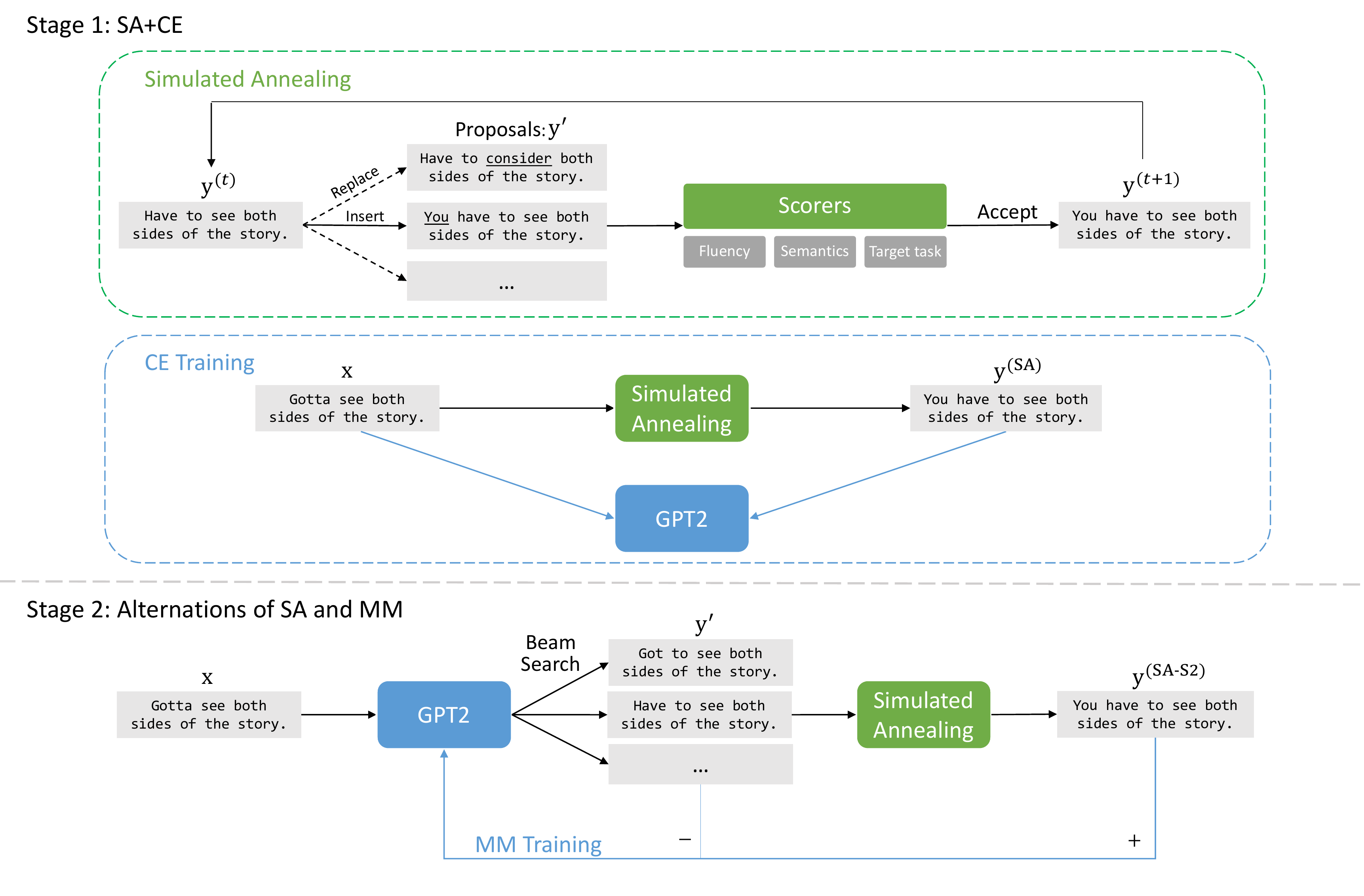}
    \caption{Two stages of search and learning in \model.}
    \label{fig:diagram}
\end{figure}
\section{Fine-Tuning Language Models for Scorers and Edit Proposal}\label{app:roberta}
Firstly, we describe the strategy of fine-tuning RoBERTa, which are used for scoring semantic compliance, and proposal of replacing words, in simulated annealing (Section~\ref{sec:SA}. In particular, we consider two fine-tuning objectives, as follows.

\textbf{Masked language model.} This fine-tuning objective is based on domain-specific unlabeled corpora, and its goal is to adapt RoBERTa more specific to the domain at hand. For each experiment, we use its unlabeled training corpus for fine-tuning.

Generally, we follow the mixed masking strategy~\cite{liu2019roberta}, which randomly masks out a few words a sentence. The goal is to predict these masked words. The mixed masking strategy randomly picks one of the three types of masking: (1) with probability 80\%, the input is substituted by a special token [MASK]; (2) with probability 10\%, the input is substituted by a random word in the vocabulary; and (3) with probability 10\%, the input is substituted by the word itself (i.e., no masking is performed).

We observe that the first mask type aligns with the \texttt{Replace} and \texttt{Insert} proposals of local editing. Thus, we would high weigh this mask type in our fine-tuning. Each time we process a data sample, we perform one more masking with the special token [MASK], in addition to the mixed strategy.  

\textbf{Formality Classification.} This objective is specific to text formalization experiment, where RoBERTa is also used for training a formality classifier. The objective is cross-entropy loss between $p(\text{formal}\,|\,\text{RoBERTa}(\text{BOS},\mathrm x)$ and the groundtruth formality label (formal or informal), where $\text{RoBERTa}(\text{BOS},\mathrm x))$ is the RoBERTa feature of a sentence $\mathrm x$ in the unlabeled dataset. Still, no parallel corpus is used.
This fine-tuning objective works together with the masked language modeling objective in a multi-task fashion.

Note that the formality classification objective does not apply to the paraphrasing task. 

As for GPT2, We first use the same strategy as pre-training to fine-tune it before search and learning. For paraphrasing, we fine-tune the model on all of the training set. For style transfer, we use the training set of formal style only.

For hyperparameters of language models fine-tuning, we performed 3 epochs of fine-tuning for text formalization and 9 epochs for paraphrasing. The maximum length of input was set to 35.  We use Adam with $\beta_1=0.9$ and $\beta_2=0.999$ for optimization. 

\section{Diagram of \model}\label{app:algo}


Figure~\ref{fig:diagram} shows the diagram of \model.
Algorithm~\ref{alg:tgsl} further presents the pseudo-code of SA search~\cite{Liu2019UnsupervisedPB} for reference. 




\section{Competing Models} \label{app:baseline}

We present more details on the competing methods in Tables~\ref{tab:auto_para_results} and~\ref{tab:auto_style_results}. All metrics are computed based on word-level tokenization (i.e., no BPE segmentation is used).

\begin{algorithm}[t]
\small
 \textbf{Input:} An input sentence $ \mathrm x$,\\
 \quad\quad\quad An initial candidate output $\rm y\brsup{0}$\\
 \textbf{Output:} An output sentence $\mathrm y$\\

 \For{$t=1, \cdots, \operatorname{MaxStep}$}{
 Set temperature $T=\max\{T_\text{init}-C\cdot t,0\}$ \quad {\color{gray}$\rhd$ $T_\text{init}, C$: annealing hyperparameters}\\
 Randomly pick an edit operator $\operatorname{Op}\in\{$ \texttt{Delete}, \texttt{Insert}, \texttt{Replace}$\}$\\
 Randomly pick an edit position $k$\\
 Propose a new candidate $\mathrm y'=\operatorname{Op}(\mathrm y\brsup{t-1}, k)$\\
 Compute acceptance rate $p(\text{accept}|\mathrm y', \mathrm{y}\brsup{t-1}, \mathrm{x}, T) = \text{min}\big\{1, \exp(\frac{s(\mathrm y'|\mathrm x)-s(\mathrm y\brsup{t-1}|\mathrm x)}{T})\big\}$\\[5pt]
 $\mathrm y\brsup{t}=\begin{cases}
 \mathrm y',&\ \ \text{with probability\ }p(\text{accept}|\mathrm y', \mathrm{y}\brsup{t-1}, \mathrm{x}, T)\\
 \mathrm y\brsup{t-1}, &\ \ \text{with probability\ } 1-p(\text{accept}|\mathrm y', \mathrm{y}\brsup{t-1}, \mathrm{x}, T)
 \end{cases}$\\
 }
 \textbf{Return:} $\mathrm y\brsup{\text{SA}}=\operatorname{argmax}_{t} s(\mathrm y\brsup{t}|\mathrm x)$
 \caption{SA for Text Generation~\cite{Liu2019UnsupervisedPB}}\label{alg:SA}
\end{algorithm}

\paragraph{Paraphrase Generation}
\begin{itemize}
    \item \textbf{RL-NN.} Qian et al.~\cite{qian2019exploring} propose to learn a reward function by neural networks and perform REINFORCE training. The results in Table~\ref{tab:auto_para_results} are from the original paper.
    \item \textbf{\textsc{Dagger}.}$^\dag$ Du et al.~\cite{du2019empirical} apply imitation learning to paraphrase generation. It achieves the state-of-the-art performance on Quora dataset. We re-ran their implementation based on our data split.
    \item \textbf{GPT2.}$^\dag$ We train another GPT2 with the same hyperparameters as our \model, but in a supervised setting for a controlled comparison.
    \item \textbf{Round-Trip MT (Transformer)}. Following~\citep{mallinson2017paraphrasing}, we utilize a well-trained bi-directional neural machine translation (NMT) system (Zh$\to$En and En$\to$Zh) with a Transformer model~\citep{vaswani2017attention}. The NMT system achieves BLEU scores of 43.2 (En$\to$Zh) and 28.74 (Zh$\to$En) on the Newstest 2017 dataset. In our work, we use the roundtrip-translation (En$\to$Zh$\to$En) as the paraphrase. 
    \item \textbf{Round-Trip MT (GPT2)}. Similarly, we adopt another GPT2-based multi-lingual (En, Zh, Es, Ru) NMT system in~\citep{guo2019zero}. Suggested by~\citep{guo2019zero}, we take the Zh as the pivot language.
    
    \item \textbf{VAE.} Variational autoencoders~\cite{bowman2015generating} generates a paraphrase by sampling from the encoded posterior distribution in the latent space. Here, we quote the results of CGMH from the implementation of~\cite{Liu2019UnsupervisedPB}.
    \item \textbf{CGMH.} Miao et al.~\cite{miao2019cgmh} propose a word-space Metropolis--Hastings approach to paraphrase generation. Results also quoted from the implementation of~\cite{Liu2019UnsupervisedPB}.
    \item \textbf{UPSA.} Liu et al.~\cite{Liu2019UnsupervisedPB} extend CGMH by decreasing the temperature and this becomes simulated annealing. Results are quoted from the original paper.
    \item \textbf{SA w/ PLM.$^\dag$} One of our extensions to UPSA is to fine-tune pretrained language models for the search objective and edit proposals. This variant is essentially the intermediate results of our \model, after its first-stage SA search.
    
\end{itemize}
While widely used for paraphrasing, the Quara dataset does not contain a standard split. The dataset is crawled from the Internet, and thus it is noisy and sometimes contains duplicate samples in training and test sets. This would not be a severe problem if the duplication is between training input and reference output during supervised learning; thus, most previous work does not explicitly deduplicate these samples. However, this could affect our \model, because we perform learning from search results with the non-parallel training set. Thus, we carefully handled this problem, ensuring no overlap in training and test.

The competing models with $\dag$ indicate that the data split is the same as \model, and the results are directly comparable. Others can be compared in a statistical sense.


\paragraph{Text Formalization}
\begin{itemize}
    \item \textbf{LSTM-attn.} Rao et al. \citep{Rao2018DearSO} trained a Bi-LSTM encoder-decoder model with attention on their parallel formality corpus.
    \item \textbf{BackTrans.} Prabhumoye et al \citep{prabhumoye2018style} utilizes back-translation to get style-independent content representations and feed them to style-dependent decoder to control the style of output.
    \item \textbf{StyleEmb.} Fu et al.~\citep{fu2018style} propose two variants for style transfer. In this variant, they accomplish style transfer by a learned style embedding.
    \item \textbf{MultiDec.} The other variant of~\citep{fu2018style} use multiple decoders for style-specific generation.
    \item \textbf{CrossAlign.} Shen et al.~\cite{shen2017style} also use style embedding, but they apply adversarial training based on style-transferred hidden states to cross-align content.
    \item \textbf{DelRetrGen.} Li et al.~\cite{li2018delete} propose a heuristic-based approach to mark style-specific words and phrases, and obtain expressions in a desired style by retrieval. Eventually, a neural model generates a style-transferred sentence.
    \item \textbf{Template.} This is a simpler variant in \citep{li2018delete}.  Then the detected style-specific words of input sentences are replaces by stylized words of target domain within its retrieved counterpart.
    \item \textbf{UnsupMT.} 
    Zhang et al.~\citep{zhang2018style} apply unsupervised machine translation techniques for style transfer. They firstly conduct word-to-word transfer and construct pseudo sentence pairs for system initialization, then conduct iterative back-translation training.
    \item \textbf{DualRL.} Luo et al.~\citep{Luo19DualRL} use a dual reinforcement learning strategy to learn bi-directional style transfer without explicitly separating the style and content. 
\end{itemize}

The results in Table~\ref{tab:auto_style_results} involve learnable metrics. We used separately trained GPT2 and RoBERT for fluency and formality evaluation, respectively. 
The GYAFC corpus has a standard dataset split. For fairness, we re-evaluated all the outputs based on our own evaluation models.

The outputs of LSTM-attn are released by \citep{Rao2018DearSO}, and the rest outputs are published by \citep{Luo19DualRL}.

 \begin{table}[!t]
    \vspace{-.2cm}
    	\small
    	\caption{Human evaluation on Quora dataset}
    	\centering
    	\label{tab:human_para_results}
    	\begin{tabular}{l|ccc}
    		\toprule
    		 \!\!\multirow{2}{*}{Method}& \!\!Grammar,\!\!\!\!  & Coherency,  & \multirow{2}{*}{Agreement} \\
    		 & Fluency & Consistency & \\
    		\midrule
    	    \!\!UPSA \citep{Liu2019UnsupervisedPB}
    		& 4.05  & 3.28  & 35.0\% \\ 
    		\!\!SA w/ PLM\!\!                    
    		& 4.79 &  4.48 & 70.0\%\\ 
    		\!\!Our \model
    		& \textbf{4.85} & \textbf{4.66} & 78.8\%\\
    		\bottomrule
    	\end{tabular}\vspace{-.3cm}
\end{table}

\ \vspace{-.4cm}

\section{Human Evaluation}\label{app:huamn}

We conduct human evaluation for the paraphrase generation experiment with selected baselines that are most relevant to our work, due to the limit of budgets.
We randomly selected 120 paraphrase samples. Four annotators with linguistic background were invited to assess each sentence pair on a scale of 1--5 with regard to two criteria: relevance (coherence \& consistent) and naturalness (grammar \& fluency) in a blind manner. Each example was reviewed by two judges and average scores are reported in Table~\ref{tab:human_para_results}. We also show the percentage of agreement between human annotators. As seen, humans have a higher agreement when the model performance is high; otherwise, the agreement is lower, because different annotators have their own scale among 1--5.    

In terms of performance, our extension of UPSA---with pretrained language models for fluency and semantic scorers---indeed improves human satisfaction in terms of Grammar/Fluency and Coherence/Consistency. Our search-and-learning \model\ model further improves the performance in both aspects.
The results are also consistent with the automatic metrics in Section~\ref{sec:auto}.
\begin{table}[t]
    \centering
        \caption{Examples generated by SA (w/ PLM) and the full \model\ model.}\label{tab:case}
        \resizebox{\linewidth}{!}{
    \begin{tabular}{cp{0.33\textwidth}p{0.33\textwidth}p{0.33\textwidth}}\toprule
         \!\!\#\!\! & Source & SA w/ PLM & \model \\ \midrule
         \!\!1\!\!& how better is sony mdr-xb950ap than sony mdr-xb450? &
         how better is the new sony mdr-xb or the sony mdr-xb? &
         how good is the sony mdr-xb950 vs the sony mdr-xb450? \\\midrule
         \!\!2\!\!&what is meant by british political conservatism? &
         what is meant by the british? &
         what is the british political conservatism?\\ \midrule
         \!\!3\!\!&what are the problems occur when flow velocity increases in a pipe? &
         what are the problems that arise when the velocity of an increase in a cylinder increases? &
         what are some problems when the flow velocity of a pipe increases? \\ \midrule
         \!\!4\!\!&if a person gets out of a vehicle can they still receive a citation? &
         can a person gets off of a vehicle, do they still receive a citation? &
         can a person get out of a car without a citation? \\  \bottomrule
    \end{tabular}}
    \label{tab:my_label}
\end{table}

\section{Case Study}\label{tab:case}
We show in Table~\ref{tab:case} examples generated by SA (with pretrained language models) and the full \model.
As seen, SA sometimes does not generate fluent sentences. In Example 2, the phrase ``\textit{political conservative}'' is deleted but no synonyms are suggested as a replacement. Our \model\ is able to generate more fluent sentences. Moreover, our \model\ generates a paraphrase in an autoregressive fashion, thus sometimes yielding a more different-appearing output, e.g., ``\textit{flow velocity increases in a pipe}'' being rephrased to ``\textit{flow velocity of a pipe increases}'' in Example 3. 

In Example 4, we also see that \model\ generates a seemingly plausible paraphrase given the input. However, the output conveys an opposite intention to the input. This shows that understanding the logic and pragmatics of language is still difficult for large-scale pretrained language models, and deeper semantic analysis shall be addressed in future work.

\end{document}